\documentclass{article}

\usepackage[preprint]{neurips_2022}

\usepackage[utf8]{inputenc} 
\usepackage[T1]{fontenc}    
\usepackage{url}            
\usepackage{booktabs}       
\usepackage{amsfonts}       
\usepackage{nicefrac}       
\usepackage{microtype}      
\usepackage{xcolor}         

\usepackage{enumitem}
\makeatletter
\newcommand{\printfnsymbol}[1]{%
  \textsuperscript{\@fnsymbol{#1}}%
}
\usepackage{algorithmicx,algpseudocode,algorithm}
\usepackage{wrapfig,tikz}
\usepackage{amsmath,longtable,fancyhdr,booktabs,multirow,graphicx,float}
\usepackage{adjustbox, bigstrut, tabularx, multirow, makecell, diagbox}
\usepackage{amssymb,xcolor,amsthm}
\usepackage{subcaption}

\newcommand{\be}{\begin{equation}}
	\newcommand{\ee}{\end{equation}}

\definecolor{Gray}{gray}{0.85}
\definecolor{LightCyan}{rgb}{0.88,1,1}

\newcolumntype{a}{>{\columncolor{Gray}}c}

\makeatletter
\usepackage{xspace}
\def\@onedot{\ifx\@let@token.\else.\null\fi\xspace}
\DeclareRobustCommand\onedot{\futurelet\@let@token\@onedot}

\title{Synthesizing PET images from High-field and Ultra-high-field MR images Using Joint Diffusion Attention Model}

\author{%
    Taofeng Xie\thanks{Taofeng Xie and Chentao Cao contributed equally to this work.}\\
    Inner Mongolia University \\
    Inner Mongolia Medical University \\
    \texttt{tf.xie@mail.imu.edu.cn} 
    \And
    Chentao Cao\printfnsymbol{1}  \\ 
    SIAT, Chinese Academy of Sciences \\ 
    \texttt{ct.cao@siat.ac.cn} 
    \AND
    Zhuo-Xu Cui  \\ 
    SIAT, Chinese Academy of Sciences \\ 
    \texttt{zx.cui@siat.ac.cn} 
    \And
    Yu Guo  \\ 
    Inner Mongolia University \\ 
    \texttt{yuguomath@aliyun.com}
    \And
    Caiying Wu  \\ 
    Inner Mongolia University \\ 
    \texttt{wucaiyingyun@163.com}
    \And
    Xuemei Wang  \\ 
    Inner Mongolia Medical University Affiliated Hospital \\ 
    \texttt{wangxuemei201010@163.com}
    \And
    Qingneng Li  \\ 
    SIAT, Chinese Academy of Sciences \\ 
    \texttt{qn.li@siat.ac.cn}
    \And
    Zhanli Hu  \\ 
    SIAT, Chinese Academy of Sciences \\ 
    \texttt{zl.hu@siat.ac.cn}
    \And
    Tao Sun  \\ 
    SIAT, Chinese Academy of Sciences \\ 
    \texttt{tao.sun@siat.ac.cn}
    \And
    Ziru Sang \\ 
    SIAT, Chinese Academy of Sciences \\ 
    \texttt{zr.sang@siat.ac.cn}
    \And
    Yihang Zhou\\ 
    SIAT, Chinese Academy of Sciences \\ 
    \texttt{yh.zhou@siat.ac.cn}
    \And
    Yanjie Zhu \\ 
    SIAT, Chinese Academy of Sciences \\ 
    \texttt{yj.zhu@siat.ac.cn}
    \And
    Dong Liang  \\ 
    SIAT, Chinese Academy of Sciences \\ 
    \texttt{dong.liang@siat.ac.cn}
    \And
    Qiyu Jin  \\ 
    Inner Mongolia University \\ 
    \texttt{qyjin2015@aliyun.com}
    \And
    Hongwu Zeng\thanks{Corresponding author.}  \\ 
    Shenzhen Children's Hospital \\ 
    \texttt{homerzeng@126.com}
    \And
    Guoqing Chen \thanks{Corresponding author.}  \\
    Inner Mongolia University \\ 
    \texttt{cgq@imu.edu.cn}
    \And
    Haifeng Wang\thanks{Corresponding author.}  \\ 
    SIAT, Chinese Academy of Sciences \\ 
    \texttt{hf.wang1@siat.ac.cn} 
}

\begin{document}

\maketitle
\begin{abstract}
MRI and PET are crucial diagnostic tools for brain diseases, as they provide complementary information on brain structure and function. However, PET scanning is costly and involves radioactive exposure, resulting in a lack of PET. Moreover, simultaneous PET and MRI at ultra-high-field are currently hardly infeasible. Ultra-high-field imaging has unquestionably proven valuable in both clinical and academic settings, especially in the field of cognitive neuroimaging. These motivate us to propose a method for synthetic PET from high-filed MRI and ultra-high-field MRI. From a statistical perspective, the joint probability distribution (JPD) is the most direct and fundamental means of portraying the correlation between PET and MRI. This paper proposes a novel joint diffusion attention model which has the joint probability distribution and attention strategy, named JDAM. JDAM has a diffusion process and a sampling process. The diffusion process involves the gradual diffusion of PET to Gaussian noise by adding Gaussian noise, while MRI remains fixed. JPD of MRI and noise-added PET was learned in the diffusion process. The sampling process is a predictor-corrector. PET images were generated from MRI by JPD of MRI and noise-added PET. The predictor is a reverse diffusion process and the corrector is Langevin dynamics. Experimental results on the public Alzheimer’s Disease Neuroimaging Initiative (ADNI) dataset demonstrate that the proposed method outperforms state-of-the-art CycleGAN for high-field MRI (3T MRI). Finally, synthetic PET images from the ultra-high-field (5T MRI and 7T MRI) be attempted, providing a possibility for ultra-high-field PET-MRI imaging.
\end{abstract}

\section{Introduction} 
Diagnosing the disease of a brain disorder (e.g., Alzheimer’s disease (AD)) has joined positron emission tomography (PET) and magnetic resonance imaging (MRI) as popular and useful methods because they can offer complementary information \cite{johnson2012brain, zhang2017pet, cheng2017cnns}. PET scanning uses radiolabeled molecules like 18F-fluorodeoxyglucose (FDG) PET to offer metabolic information. Meanwhile, MRI scanning can provide structural information. However, due to the expense of PET scanning and the potential risks associated with radioactive exposure, some patients may decline it, resulting in a lack of PET. Simultaneous PET-MRI scans at ultra-high-field are not feasible with the existing technology. Ultra-high-field imaging has unquestionably proven valuable in both clinical and academic settings, especially in the field of cognitive neuroimaging. The signal-to-noise ratio (SNR) of the ultra-high-field MRI increases and can provide more pathophysiology data. 
Similarly, the functional identification of specific cortical areas was improved by combining the high resolution of MRI with the high molecular imaging capabilities of PET to achieve an accuracy that can be compared to the intricate brain architecture \cite{cho2008fusion}. This knowledge may aid in the earlier detection and treatment of cerebrovascular disorders, brain tumours, ageing-related alterations, and multiple sclerosis \cite{van2013clinical,beisteiner2011clinical}. 
However, PET, which corresponds to MRI in the ultra-high field, cannot be achieved. The inspiration for the synthesis of high-resolution PET was the exceptional performance of the ultra-high-field MRI. 
For the above two reasons, missing PET imaging be synthesized urgently in order to make effective use of the multi-modality medical data. The next question to consider is whether the ultra-high-field strength has an effect on PET imaging.

The effect of a magnetic field on positron emitters has been studied in many papers. The MRI field improves the resolution and contrast of PET images \cite{shah2014effects}. A system of fusion between a high-resolution PET and an ultra-high-field MRI became possible \cite{cho2008fusion}. The paper discussed the hybrid PET-MRI technology and the ultra-high-field together with a PET insert that has been feasible \cite{shah2013advances}. MRI on the different B0 had negligible effects on PET imaging by measuring the effects of MRI on the flood histograms and energy resolutions of PET \cite{sang2022mutual}. 

The synthetic PET by the high-field MRI or the ultra-high-field MRI is a problem of cross-modal generation \cite{hussein2022brain, taofeng2022brain}.
Initially, the generated model learned the mapping between the real image and the target image by extracting features \cite{yu2020medical}. Then, this approach was limited to manually designed feature representations. 
In recent years, deep learning has had great success in medical imaging. Deep learning has stronger feature representations in a data-driven manner. Generative model using deep neural networks has been a popular approach. Synthesis of $^{15}O-water$ PET CBF from multi-sequence MRI by an encoder-decoder network with attention mechanisms \cite{hussein2022brain}. Synthesis of PET from corresponding MRI used an effective U-Net architecture with the adversarial training strategy \cite{hu2019cross}. 
From the perspective of probabilistic modelling, the key to generative models is that the distribution of generated samples is the same as the distribution of the real data. 
Generative models have variational autoencoders (VAEs) \cite{rezende2014stochastic, oussidi2018deep}, normalizing flows \cite{dinh2016density}, generative adversarial networks (GAN) \cite{goodfellow2020generative} and diffusion model \cite{sohl2015deep} and so on. The samples generated by VAEs are often unrealistic and ambiguous due to the way in which the data distribution is estimated and the loss function is computed \cite{kingma2013auto}. 
Normalizing flow converts a simple distribution into a complex distribution by applying a series of reversible transformation functions, where the variable substitution theorem can be used to obtain the expected probability distribution of the final target variable \cite{kobyzev2020normalizing}. 
GANs are state-of-the-art image generation tasks based on sample quality measures \cite{brock2018large, karras2020analyzing, wu2019logan}. 
GANs can generate high-quality samples, but can't do sampling diversity \cite{nash2021generating, razavi2019generating}. GANs are unstable in the learning process due to pattern collapse during training \cite{miyato2018spectral}. These issues may restrict the variety and quality of generated images, especially in medical imaging. Diffusion models, however, show promise as a substitute for GANs, as they can generate sufficient pattern coverage and high-quality sampling, addressing the limitations of other generative models.

The diffusion model is a class of likelihood-based models that have gained widespread interest as a generative model in the field of medical imaging. The tasks of the diffusion model have image generation \cite{dhariwal2021diffusion}, image super-resolution \cite{li2022srdiff}, image inpainting \cite{lugmayr2022repaint}, image segmentation \cite{amit2021segdiff} and so on. 
In particular, the diffusion model has remarkable advances beat GANs \cite{dhariwal2021diffusion}. Recently, the application of diffusion models in medical imaging has grown exponentially based on the score matching with Langevin dynamics (SMLD) and the denoising diffusion probabilistic modeling (DDPM). SMLD \cite{song2019generative} calculates the score (i.e., the gradient of the log probability density with respect to the data) at each noise scale and samples from a series of diminishing noise scales during generation using Langevin dynamics. 
DDPM \cite{ho2020denoising} uses functional knowledge of the reverse distributions to make training tractable. It trains a series of probabilistic models to reverse each stage of noise corruption. 
The model of score-based generative solving the stochastic differential equation (SDE) \cite{song2020score} unified framework of SMLD \cite{song2019generative} and DDPM \cite{ho2020denoising}. SDE diffuses a data point into random noise and then reverses the process to mold random noise into raw data continuously. Currently, SDE can transfer a Gaussian noise to an image distribution. However, it cannot transfer from a modal distribution to the other modal distribution.

Given the shortcoming of the diffusion model, the joint diffusion attention model (JDAM) was constructed for synthetic PET from MRI.
In other words, PET which is not easily available was generated from MRI which is easily available by JPD of MRI and PET with noise. The JDP is the most straightforward and fundamental way to depict the connection between PET and MRI from a statistical standpoint. JDAM has a diffusion process and a sampling process. The diffusion process is that PET diffuses to Gaussian noise gradually by adding Gaussian noise, and MRI is fixed. JPD of MRI and PET with noise was learned in the diffusion process. The sampling process is a predictor-corrector. The predictor and corrector are executed alternately. The predictor is a reverse diffusion process and the corrector is Langevin dynamics. On the public Alzheimer’s Disease Neuroimaging Initiative (ADNI) dataset, we contrasted CycleGAN, SDE, and our model achieved a great result in the high-field. And we extend JDAM from the high-field to the ultra-high-field.

The rest of this paper is organized as follows. Section \ref{Sec:2} introduces the backwardness of the diffusion model. Section \ref{Sec:3} describes materials and methods. Section \ref{Sec:4} provides experimental results. Discussion is given in Section \ref{Sec:5}. The conclusion is presented in Section \ref{Sec:6}.

\section{Backward}
\label{Sec:2}
A diffusion process $\left\{x(t)\right\}_{t=0}^{T}$ is constructed, where $t\in{\left[0, T\right]}$ is a continuous variable of the time index of the process. $x(t)\sim p_{t}(x)$ denote that $p_{t}(x)$ is the probability density of $x_{t}$, where $x(t)$ is denoted as $x_{t}$. $x_{0}$ denotes an original image, while $x_{T}$ is the outcome by adding noise to $x_{0}$ gradually.
Let $x_{0}\sim p_{data}$, where $p_{data}$ is an unknown data distribution, and $x_{T} \sim p_{T}$, where $p_{T}$ represents a prior distribution that excludes information about $p_{data}$. 
The diffusion process is described as the solution of the following SDE:
$$\mathrm{d}x=f(x,t)\mathrm{d}t+g(t)\mathrm{d}w,$$
where $f$ is the drift coefficient of $x_{t}$, $g$ is the diffusion coefficient, and $w$ is a standard Brownian motion. 
The reverse of the diffusion process, that is $x_{T}$ diffusion to $x_{0}$, can be described as a solution of the reverse-time SDE
$$\mathrm{d}x=\left[f(x,t)-g(t)^{2}\nabla_{x} \log p_{t}(x)\right]\mathrm{d}t+g(t)\mathrm{d}\bar{w},$$
where $\nabla_{x}\log p_{t}(x)$ is the score of $p_{t}(x)$ and is estimated by the score network $s_{\theta}\left(x(t),t\right)$. 
$\nabla_{x}\log p_{t}(x)$ was replaced by $\nabla_{x}\log p_{0t}(x(t) \mid x(0))$ because $\nabla_{x}\log p_{t}(x)$ is unknown. 
Where $p_{0t}\left(x(t)\mid x(0)\right)$ is the Gaussian perturbation kernel that perturbs the probability density $p_{data}(x)$ to $p_{t}(x)$ by Gaussian noise.
Let $p_{\sigma}(\bar{x}\mid x):=\mathcal{N}\left(\bar{x};x,\sigma^{2}I\right)$ is a perturbation kernel, $p_{\sigma}\left(\bar{x}\right):=\int p_{data}(x)p_{\sigma}\left(\bar{x}\mid x\right)\mathrm{d}x$.
The parameters $\theta$ of the score network are optimized using the optimization problem. 
$$\boldsymbol{\theta}^{*}=\underset{\boldsymbol{\theta}}{\arg \min}\mathbb{E}_{t}\left\{\lambda(t)\mathbb{E}_{x(0)}\mathbb{E}_{x(t)|x(0)}\left[\left\| s_{\theta}\left(x(t),t\right)-\nabla_{x(t)}\log p_{0t}\left(x(t)\mid  x(0)\right)\right\|_{2}^{2}\right]\right\}$$
Where $\lambda(t)$ is a reasonable weighting function.

\section{Materials and methods}
\label{Sec:3}
\subsection{Data acquisition and preprocessing}
The study utilized the Alzheimer’s Disease Neuroimaging Initiative (ADNI) dataset \cite{jack2008alzheimer}. ADNI comprises T1-weighted structural MRI, fluorodeoxyglucose (FDG) PET imaging, and other imaging modalities. In total, 13560 pairs of images from 452 subjects were used for our experiments. 
For data preprocessing, MRI was processed in five steps: anterior commissure (AC) and posterior commissure (PC) (AC-PC) alignment, skull stripping, intensity correction, cerebellum removal, and spatial alignment to the same standardized Montreal Neurological Institute (MNI) coordinate space.

Firstly, AC-PC alignment by Statistical Parametric Mapping (SPM)\footnote{https://www.fil.ion.ucl.ac.uk/spm}. 
Secondly, non-brain tissue was eliminated from MRI by HD-BET \cite{isensee2019automated} to avoid interference from redundant information. 
None-brain tissue was also eliminated from PET using the mask of the non-brain tissue in MRI. Thirdly, the MRI and PET of each subject were spatially aligned, with each PET aligned to its corresponding MRI by registration. MRI of each subject was spatially aligned to the same standardized Montreal Neurological Institute (MNI) coordinate space and PET of the same subject was aligned to the MRI using FMRIB's Software Library (FSL)\footnote{https://fsl.fmrib.ox.ac.uk/fsl}.
Moreover, MRI and PET images were reshaped to $109 \times 109 \times 91$ voxels. 
Each MRI and PET volume consisted of a 2D axial image slice and was resampled to $128 \times 128$ for each slice of MRI and PET. 
Our experiment takes the 30th slice to the 60th slice from each subject of axial, coronal, and sagittal planes respectively.

\subsection{Joint diffusion attention model}
\begin{figure}[t]
    \centering
    \includegraphics[width=1\textwidth]{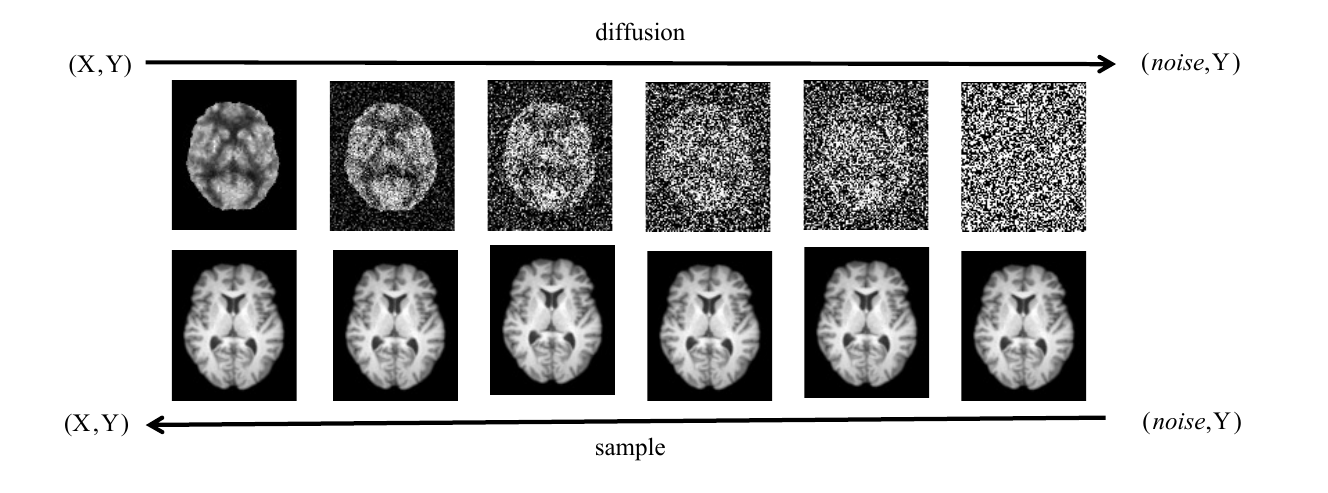}
    \caption{Schematic diagram of the joint diffusion attention model. }
    \label{fig: schematic diagram}
\end{figure}

Diffusion modeling is a crucial method for generating models. This method forecasts the score,  which is the gradient of the log probability density with respect to the original data, rather than the data distribution directly. 
The original diffusion model recovers the data from the noise by reversing the process that destroys data through additional noise, resulting in a generative image from noise. However, generating accurate medical images from noise is not guaranteed for medical images. This paper provides conditions as guidance to achieve the expected results, such as synthetic PET by MRI as a guide. Given co-registered FDG PET and T1-weighted MRI pairs $\left(\mathbf{x},\mathbf{y}\right)$. 
$\mathbf{x}$ denotes the data of PET and $\mathbf{y}$ represents the data of MRI. 
The conditional probability distribution $p\left(\mathbf{x} \mid \mathbf{y}\right)$ needs to be estimated. However, it is not easy to estimate. $\nabla_{\mathbf{x}}p\left(\mathbf{x}\mid \mathbf{y}\right)$ can be considered. In the framework of diffusion models, $\nabla_{\mathbf{x}}p\left(\mathbf{x}\mid \mathbf{y}\right)$ needs to be estimated. Due to $\nabla_{\mathbf{x}}p\left(\mathbf{x}\mid \mathbf{y}\right)=\nabla_{\mathbf{x}}p\left(\mathbf{x}, \mathbf{y}\right)$, we will construct an MRI and PET joint diffusion model and learn its joint distribution to achieve conditional generation. The joint diffusion attention model (JDAM) has two processes including the diffusion process and the sampling process.
Figure \ref{fig: schematic diagram} is a schematic diagram of the joint diffusion attention model.

The diffusion process of JDAM is performed only in PET, and MRI is fixed. 
The initial value of the reverse diffusion is the MRI image plus the Gaussian noise. 
Based on the framework of the SDE, given a forward SDE
\begin{equation*}
    \mathrm{d}\mathbf{x}=f(\mathbf{x},t)+g(t)\mathrm{d}\mathbf{w},
    \label{SDE}
\end{equation*}
where $f_{i}$ denotes the drift coefficient of $\mathbf{x}_{i}$ and $g_{i}$ denotes the diffusion coefficient. Here, $\mathbf{w}$ is a standard Brownian motion. 
The discrete form can be described as 
\begin{equation*}
\mathbf{x}_{i+1}=\mathbf{x}_{i}+f_{i}(\mathbf{x}_{i})+g_{i}z_{i},~i=0,1,...,N, 
\label{discrete_SDE}
\end{equation*}
where $z_{i}\sim \mathcal{N}\left(0,I\right)$, and $\mathbf{x}$ add noise gradually while $\mathbf{y}$ is fixed in a forward SDE. 
The reverse-time SDE can be expressed as
\begin{equation*}
\mathrm{d}\mathbf{x}=\left[f(\mathbf{x},t)-g(t)g(t)^{T}\nabla_{\mathbf{x}}\log p_{t}\left(\mathbf{x}\mid \mathbf{y}\right)\right]\mathrm{d}t+g(t)\mathrm{d}\mathbf{w}.
\end{equation*}
According to the properties of logarithm and derivative
\begin{equation*}
\begin{aligned}
\nabla_{\mathbf{x}}\log p_{t}(\mathbf{x}\mid \mathbf{y}) 
&=\nabla_{\mathbf{x}}(\log p_{t}\left(\mathbf{x},\mathbf{y}\right) - \log p_{t}\left(\mathbf{y})\right) \\
&=\nabla_{\mathbf{x}}\log p_{t}\left(\mathbf{x},\mathbf{y}\right),
\end{aligned}
\end{equation*}
the reverse-time SDE can be rewritten as
\begin{equation*}
\mathrm{d}\mathbf{x}=\left[f(\mathbf{x},t)-g(t)g(t)^{T}\nabla_{\mathbf{x}}\log p_{t}\left(\mathbf{x}, \mathbf{y}\right)\right]\mathrm{d}t+g(t)\mathrm{d}\mathbf{w}.
\end{equation*}
As a similar functional form, the iteration rule be written as 
\begin{equation*}
\mathbf{x}_{i}=\mathbf{x}_{i+1}-f_{i+1}(\mathbf{x}_{i+1})+g_{i+1}g_{i+1}^{T}s_{\theta^{*}}\left(\mathbf{x}_{i+1},\mathbf{y},i+1\right)+g_{i+1}z_{i+1},
\end{equation*}
where $s_{\theta^{*}}(\mathbf{x}_{i+1},\mathbf{y},i+1)$ is to estimate $\nabla_{\mathbf{x}_{i+1}}\log p_{t}(\mathbf{x}_{i+1},\mathbf{y})$, and $p_{t}(\mathbf{x}_{i+1},\mathbf{y})$ is the joint probability distribution of $\mathbf{x}_{i+1}$ and $\mathbf{y}$. $s_{\theta^{*}}(\mathbf{x}_{i+1},\mathbf{y},i+1)$ is obtained by U-Net. the U-Net be described as the optimization problem
\begin{equation*}
    \boldsymbol{\theta}^{*}=\underset{\boldsymbol{\theta}}{\arg \min } \mathbb{E}_{t}\left\{\lambda(t) \mathbb{E}_{\mathbf{x}(0)} \mathbb{E}_{\mathbf{x}(t) \mid \mathbf{x}(0)}\left[\left\|s_{\boldsymbol{\theta}}\left(\mathbf{x}(t), \mathbf{y}, t\right)-\nabla_{\mathbf{x}(t)} \log p_{0 t}\left(\mathbf{x}(t) \mid \mathbf{x}(0)\right)\right\|_{2}^{2}\right]\right\},
    \label{SDE-loss}
\end{equation*}
where the gradient of the perturbation kernel is
\begin{equation*}
\nabla_{\mathbf{x}(t)} \log p_{0t}\left(\mathbf{x}_{t}\mid \mathbf{x}_{0}\right)=-\frac{\mathbf{x}(t)-\mathbf{x}(0)}{\sigma^{2}(t)}.
\end{equation*}
The loss function be transformed as
\begin{equation*}
    \boldsymbol{L}^{*}= \mathbb{E}_{t}\left\{\lambda(t) \mathbb{E}_{\mathbf{x}(0)} \mathbb{E}_{\mathbf{x}(t) \mid \mathbf{x}(0)}\left[ \left\|\sigma(t)s_{\boldsymbol{\theta}}\left(\mathbf{x}(t), \mathbf{y}, t\right)+\frac{\mathbf{x}(t)-\mathbf{x}(0)}{\sigma(t)}\right\|_{2}^{2}\right]\right\},
    \label{SDE-loss_N}
\end{equation*}
where $\lambda(t)$ is a positive weighting function, and $\frac{\mathbf{x}(t)-\mathbf{x}(0)}{\sigma(t)}\sim \mathcal{N}(0,I)$. 

\begin{figure}[t]
    \centering
    \includegraphics[width=1\textwidth]{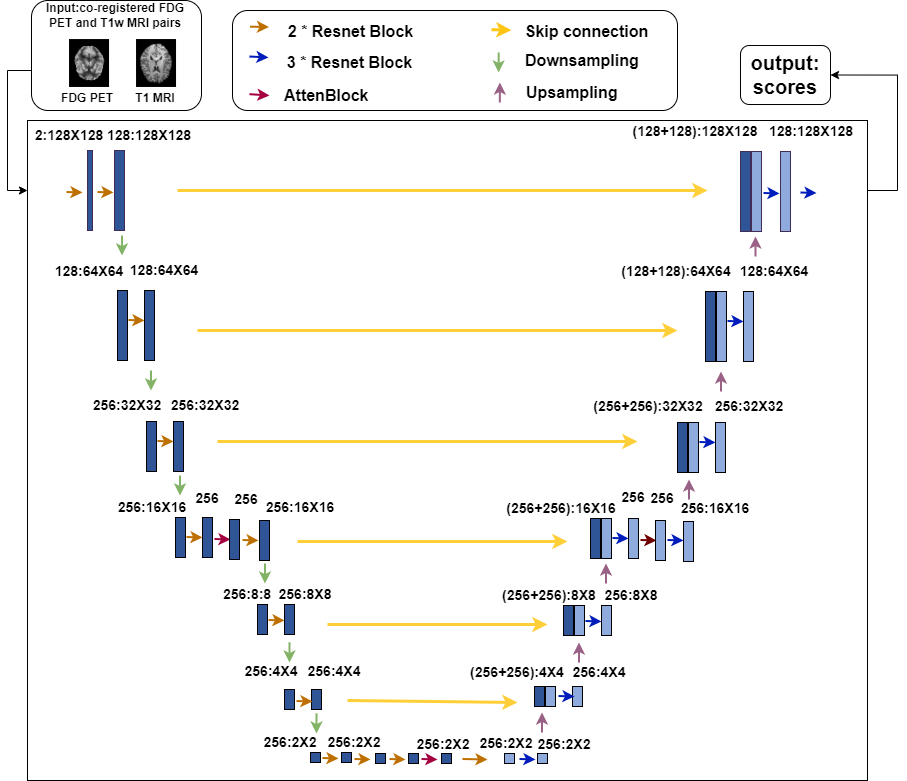}
    \caption{Overview of the proposed U-Net architecture incorporating Resnet block and Attention block into the U-Net.}
    \label{fig:UNet}
\end{figure}

\begin{figure}[t]
    \centering
    \includegraphics[scale=0.5]{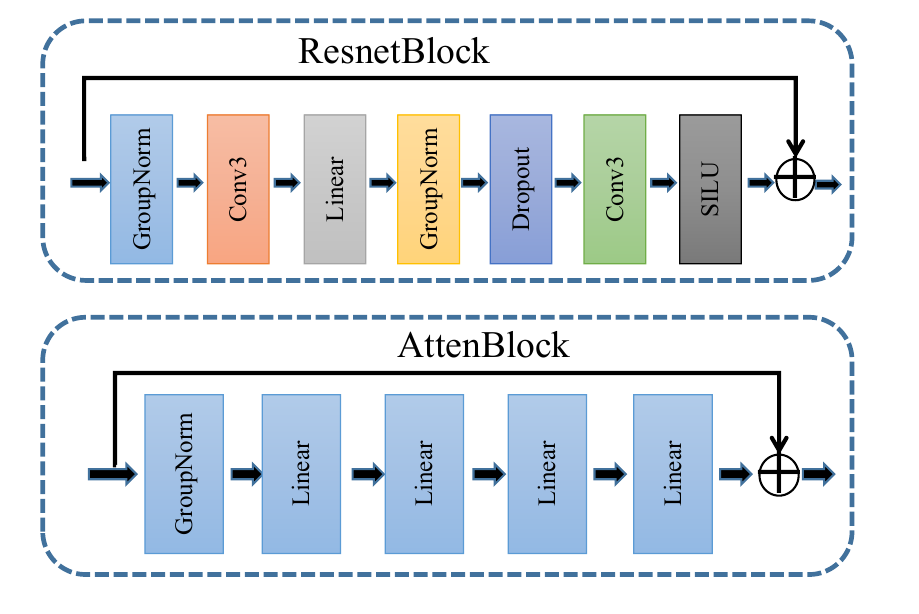}
    \caption{The architecture of a Resnet block and an Attention block used in the U-Net.}
    \label{fig:ResnetBlock}
\end{figure}

Figure \ref{fig:UNet} shows the architecture of the U-Net. The input to the network is a 2-channel tensor that includes MRI and PET. The output of the network is a score $s_{\theta}(\mathbf{x}, \mathbf{y},t)$ that is the gradient of the log joint probability density, where $\theta$ contains the network parameters to be learned. The input images were passed through the encoder where the first three blocks are Resnet blocks, the next block is the Resnet block add Attention block, and the last three blocks are Resnet blocks. The decoder has three Resnet blocks, the Resnet block adds an Attention block followed by three Resnet blocks. The Resnet block and the Attention block be shown in Figure \ref{fig:ResnetBlock}. Skip connections are also included between the encoder and decoder. Each Resnet block is equipped with a GroupNorm layer, a convolution layer, a linear layer, a GropuNorm layer, a dropout layer, a convolution layer, and a SILU layer. The Attention block consists of a GroupNorm layer and four linear layers.

SDE \cite{song2020score} proposed three different forms of SDEs, (i.e., VE-SDE, VP-SDE, and sub-VP-SDE). In this work, we choose the VE-SDE. The VE-SDE can be regarded as a continuous form of the noise perturbations SMLD noise perturbation. The perturbation kernel of SMLD can be described as 
\begin{equation*}
    p_{\sigma}\left(\mathbf{x}\mid \mathbf{x}_{0}\right)\sim \mathcal{N}\left(\mathbf{x};\mathbf{x}_{0},\sigma^{2}\mathbf{I}\right).
\end{equation*}
Corresponding distribution of $\mathbf{x}_{i}$ satisfies the Markov chain
\begin{equation}
    \mathbf{x}_{i}=\mathbf{x}_{i-1}+\sqrt{\sigma_{i}^{2}-\sigma_{i-1}^{2}}z_{i-1},~i=1,2,...,N,
    \label{Markov chain}
\end{equation}
The noise scales $\lbrace{\sigma_{i}\rbrace}_{i=1}^{N}$ satisfies that $\sigma_{1}$ is large sufficiently, i.e., $\sigma_{max}$. $\sigma_{N}$ is small enough, i.e., $\sigma_{min}$. 
Let $ N \rightarrow \infty $, $\lbrace{\mathbf{x}_{i}\rbrace}_{i=1}^{N}$ becomes a continue function $\mathbf{x}(t)$, and $\lbrace{\sigma_{i}\rbrace}_{i=1}^{N}$ becomes $\sigma(t)$, and $\lbrace{z_{i}\rbrace}$ becomes $ z(t) $. 
A continuous time variable $t\in [0,1]$ is introduced. 
The Markov chain \eqref{Markov chain} can be rewritten in continuous form as 
\begin{equation}
    \mathbf{x}(t+\Delta t)=\mathbf{x}(t)+\sqrt{\sigma^{2}(t+\Delta t)-\sigma^{2}(t)}z(t)\approx \mathbf{x}(t)+\sqrt{\frac{\mathrm{d}[\sigma^{2}(t)]}{\mathrm{d}t}\Delta t}z(t),
    \label{continue}
\end{equation}
where $\Delta t \ll 1$ ensures that the approximation equality holds. 
In the limit of $\Delta t \to 0$, \eqref{continue} converges to 
\begin{equation*}
    \mathrm{d}\mathbf{x}=\sqrt{\frac{\mathrm{d}[\sigma^{2}(t)]}{\mathrm{d}t}}\mathrm{d}\mathbf{w}.
\end{equation*}

Since $\lbrace{\sigma_{i}\rbrace}_{i=1}^{N}$ is a geometric sequence, let $\sigma_{i}=\sigma(\frac{i}{N})=\sigma_{min}\left(\frac{\sigma_{max}}{\sigma{min}}\right)^{\frac{i-1}{N-1}}$ for $i=1,2,...,N$. 
In the limit of $N \rightarrow \infty$, we have $\sigma(t)=\sigma_{min}\left(\frac{\sigma_{max}}{\sigma{min}}\right)^{t}$ for $t\in (0,1]$. The corresponding VE-SDE is given by
\begin{equation*}
\mathrm{d}\mathbf{x}=\sigma_{min}\left(\frac{\sigma_{max}}{\sigma_{min}}\right)^{t}\sqrt{2 \log \frac{\sigma_{max}}{\sigma_{min}}}\mathrm{d}\mathbf{w},t\in (0,1].
\end{equation*}
The perturbation kernel can be rewritten as
\begin{equation*}
    p_{0t}\left(\mathbf{x}(t)\mid \mathbf{x}(0)\right)=\mathcal{N}\left(\mathbf{x};\mathbf{x}_{0},\sigma_{min}^{2}\left(\frac{\sigma_{max}}{\sigma_{min}}\right)^{2t}\mathbf{I}\right).
\end{equation*}
The diffusion process is described as follows:
\begin{equation*}
\mathbf{x}_{i+1} = \mathbf{x}_{i} + \sigma_{min} \left( \frac{\sigma_{max}}{\sigma_{min}} \right) ^{i} z, ~ i=1,2,...,N-1,
\end{equation*} 
where $\mathbf{x}_{i}$ is the $i$-th  perturbed data of PET, $\mathbf{x}_{0}$ is the data of PET, and $\mathbf{x}_{T}$ is prior information of the PET that obeys Standard Gaussian distribution. 
The set of positive noise scales is $\{\sigma_{i}\}_{i=1}^{N}$, $\sigma_{min}$ is the minimum of the noise scales satisfying $p_{\sigma_{min}}(x)=p_{data}(x)$, and $\sigma_{max}$ is the maximum of the noise scales. 
 
 \begin{algorithm}[t]
	\caption{Training.}
	\label{alg:training}
	\begin{algorithmic}[1]
 \Require{$ \{\sigma_i\}_{i=1}^N$.}
              \State{$(\mathbf{x}_{0},\mathbf{y})\sim p(\mathbf{x},\mathbf{y})$}
          \State{$t\sim \mathcal{U}(0,1)$}
          \State{$z\sim \mathcal{N}(0,I)$}
          \State{$\mathbf{x}_{t}=\mathbf{x}_{0}+\sigma_{min} \left( \frac{\sigma_{max}}{\sigma_{min}} \right) ^{i}z$}
          \State{$\underset{\theta}{\mathrm{min}}\left\| \sigma_{min} \left( \frac{\sigma_{max}}{\sigma_{min}} \right) ^{i}s_{\theta}\left(\mathbf{x}(t), \mathbf{y}, t\right)+z \right\|^{2}$}
	   \State{$\mathbf{x}_{N} \sim \mathcal{N}(0, 
          \sigma(T) I$)}
        \State{until converged}
	\end{algorithmic}
\end{algorithm}
The training process is shown in Algorithm \ref{alg:training}.
\begin{algorithm}[t]
	\caption{PC Sampling.}
	\label{alg:PC Sampling}
	\begin{algorithmic}[1]
	    \Require{$ \{\sigma_i\}_{i=1}^N, \mathbf{y}, r, N, M$}
	   \State{$\mathbf{x}_{N} \sim \mathcal{N}\left(0, \sigma(N) I\right)$}
	    \For{$i = N-1$ to $0$}
	        \State{$ z \sim \mathcal{N}(0, I)$}
	        \State{$\mathbf{x}_{i} \leftarrow \mathbf{x}_{i+1}+\left(\sigma_{i+1}^{2}-\sigma_{i}^{2}\right)s_{\boldsymbol{\theta^{*}}}\left(\mathbf{x}_{i+1},\mathbf{y}, \sigma_{i+1}\right)+\sqrt{\sigma_{i+1}^{2}-\sigma_{i}^{2}}z$}
            \For{$j \gets 1$ to $M$}
                \State{$z \sim \mathcal{N}\left(0, I\right)$}
                \State{$\epsilon \leftarrow 2 \left(r \frac{\|z\|_{2}}{\left\|s_{\theta^{*}}\left(\mathbf{x}_{i}^{j-1},\mathbf{y}, i\right)\right \|_{2}} \right)^{2}$}                
                \State{$\mathbf{x}_{i}^{j} \leftarrow \mathbf{x}_{i}^{j-1}+\epsilon s_{\theta^{*}}\left(\mathbf{x}_{i}^{j-1},\mathbf{y}, i\right)+\sqrt{2 \epsilon} z$}
            \EndFor
            \State{$\mathbf{x}_{i-1}^{0} \leftarrow \mathbf{x}_{i}^{M}$}
        \EndFor
        \item[]
        \Return{${\bf x}_0^0$}
	\end{algorithmic}
\end{algorithm}

The sampling process is a predictor-corrector called a PC sample. The predictor and corrector are executed alternately. The predictor is a reverse diffusion process.
\begin{equation*}
\mathbf{x}_{i} = \mathbf{x}_{i+1} - f_{i+1}(\mathbf{x}_{i+1}) + g_{i+1}g_{i+1}^{T} s_{\theta^{*}} \left(\mathbf{x}_{i+1},\mathbf{y},i+1 \right) + g_{i+1}z_{i+1}.
\end{equation*}
 The corrector is Langevin dynamics, which can compute the sample by
\begin{equation*}
\mathbf{x}_{i} = \mathbf{x}_{i+1} + \varepsilon s_{\theta}\left(\mathbf{x}_{i+1},\mathbf{y},i+1\right) + \sqrt{2\varepsilon} z.
\end{equation*}
Here, $\varepsilon=2 \alpha_{i}\left(r\lVert z\rVert_{2} / \lVert s_{\theta}\rVert_{2}\right)$ denotes step size, where $r$ is the "signal-to-noise" ratio.
Therefore, the sampling process can be summarized in Algorithm \ref{alg:PC Sampling}.

\section{Experimental results}
\label{Sec:4}

\begin{figure}[t]
    \centering
    \includegraphics[width=1\textwidth]{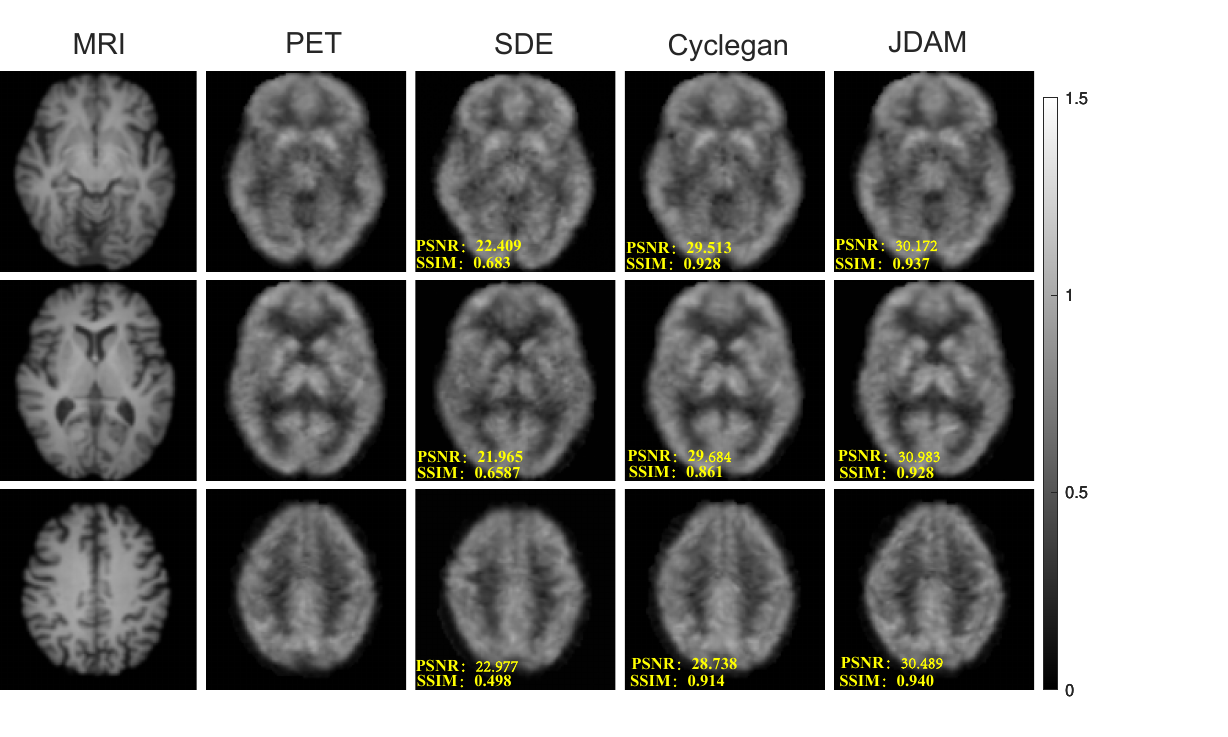}
    \caption{Comparison of synthetic results in the axial plane. The first row shows MRI, corresponding ground truth PET and synthetic PET results obtained using score-based SDE, CycleGAN, and JDAM methods in the upper layer of the axial plane. The second row shows images whose order is the same as the first row in the middle layer of the axial plane, while the third row shows images whose order is the same as the first row in the lower layer. The yellow numbers in the bottom left corner indicate the PSNR (dB) and SSIM between real PET and synthesis PET.}
    \label{fig:x}
\end{figure}
\begin{figure}[t]
    \centering
    \includegraphics[width=1\textwidth]{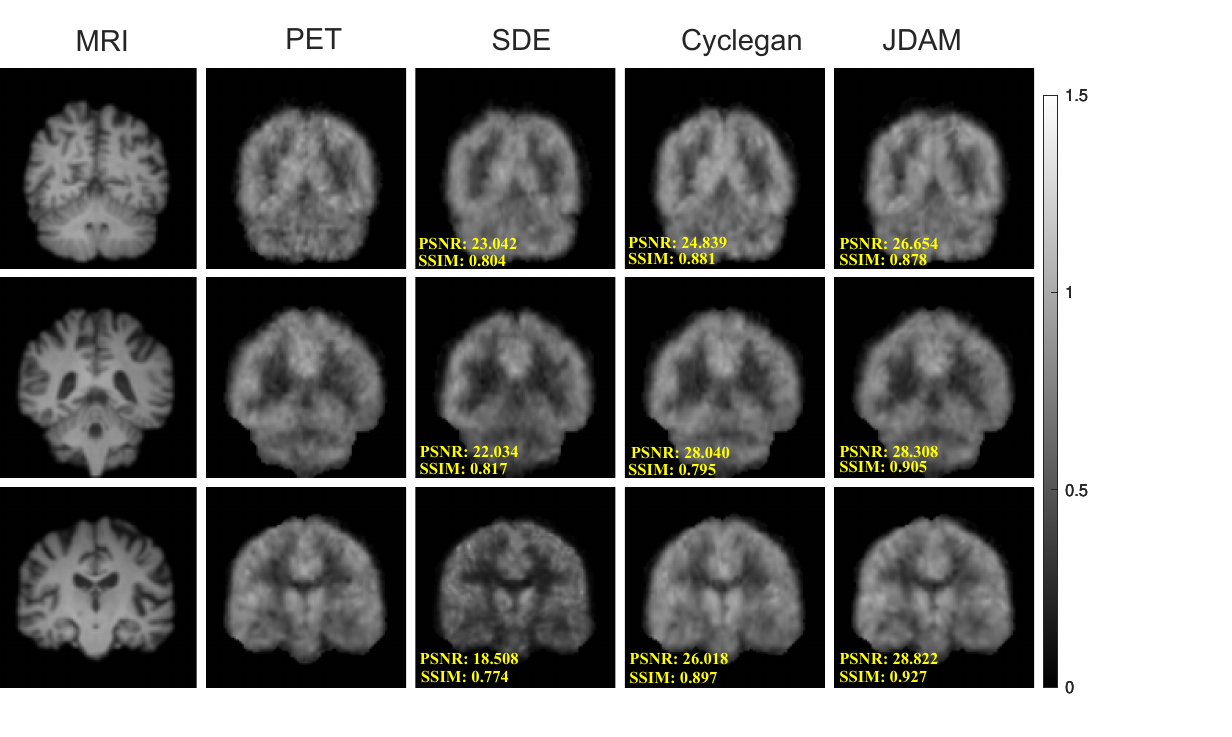}
    \caption{Comparison of synthetic results in the coronal plane. The first row shows MRI, corresponding ground truth PET and synthetic PET results obtained using score-based SDE, CycleGAN, and JDAM methods in the upper layer of the coronal plane. The second row shows images whose order is the same as the first row in the middle layer of the coronal plane, while the third row shows images whose order is the same as the first row in the lower layer. The yellow numbers in the bottom left corner indicate the PSNR (dB) and SSIM between real PET and synthesis PET.}
    \label{fig:y}
\end{figure}
 \begin{figure}[t]
    \centering
    \includegraphics[width=1\textwidth]{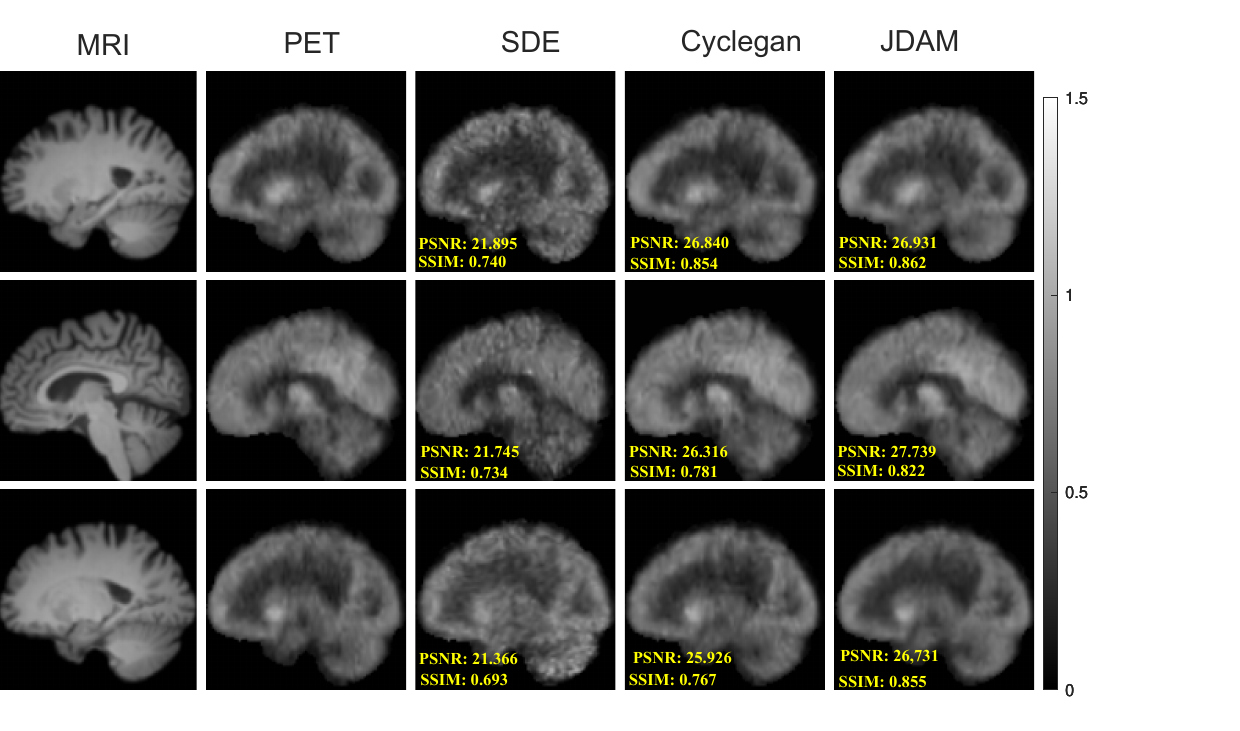}

    \caption{Comparison of synthetic results in the sagittal plane. The first row shows MRI, corresponding ground truth PET and synthetic PET results obtained using SDE, CycleGAN, and JDAM methods in the upper layer of the sagittal plane. The second row shows images whose order is the same as the first row in the middle layer of the sagittal plane. The third row shows images whose order is the same as the first row in the lower layer. The yellow numbers in the bottom left corner indicate the PSNR (dB) and SSIM between real PET and synthesis PET.}
    \label{fig:z}
\end{figure}

The experimental results are compared against SDE and CycleGAN to support the performance of JDAM. SDE can be seen as an ablation experiment because SDE indicates the generation of PET images from noise. The ablation experiments prove the advantages of JPD. Meanwhile, CycleGAN is a state-of-the-art generative model.
This study uses peak signal-to-noise ratio (PSNR) and structural similarity (SSIM) for quantitative assessment, where PSNR assesses the error, and SSIM reflects structural similarities between real PET images and generative PET images. The qualitative observations of three samples in the axial plane, coronal plane, and sagittal plane are demonstrated in Figures \ref{fig:x}, \ref{fig:y}, and \ref{fig:z}.
The results show that the PET images generated by JDAM are the most similar to the real PET images. It has been established that JDAM performs superiorly to other methods in the task of synthesizing PET using MRI as a guide. 
Figure \ref{fig:x} shows the generated samples for different models. 
Compared to SDE and CycleGAN, our generated results are visually pleasing and contain better details. 
The good visual quality of the generated samples explains that our method is feasible for generative medical images. Figure \ref{fig:x} shows examples of MRI, true PET, and synthesized PET images obtained by the SDE, CycleGAN, and JDAM, across three layers, including the upper layer, the media layer, and the lower layer. PSNR and SSIM were computed to evaluate all methods. The PSNR and SSIM values of the JDAM are the highest.
Notably, the structure of synthetic PET by SDE is different from that of real PET.

Figure \ref{fig:y} and Figure \ref{fig:z} show the visualizations of synthetic PET for the same representative subjects along the coronal plane and sagittal plane, respectively. 
The results of the coronal plane and sagittal plane were similar to those of the axial plane. 
The synthesis results of JDAM are better than those of other methods in the high field. 
The efficiency of the JDAM is better. The PSNR and SSIM for JDAM are higher than for other methods. 
To further illustrate the effectiveness of JDAM, we compare the quality of synthetic and real PET for healthy controls and Alzheimer's disease patients.
\begin{figure}
    \centering
    \includegraphics[width=1\textwidth]{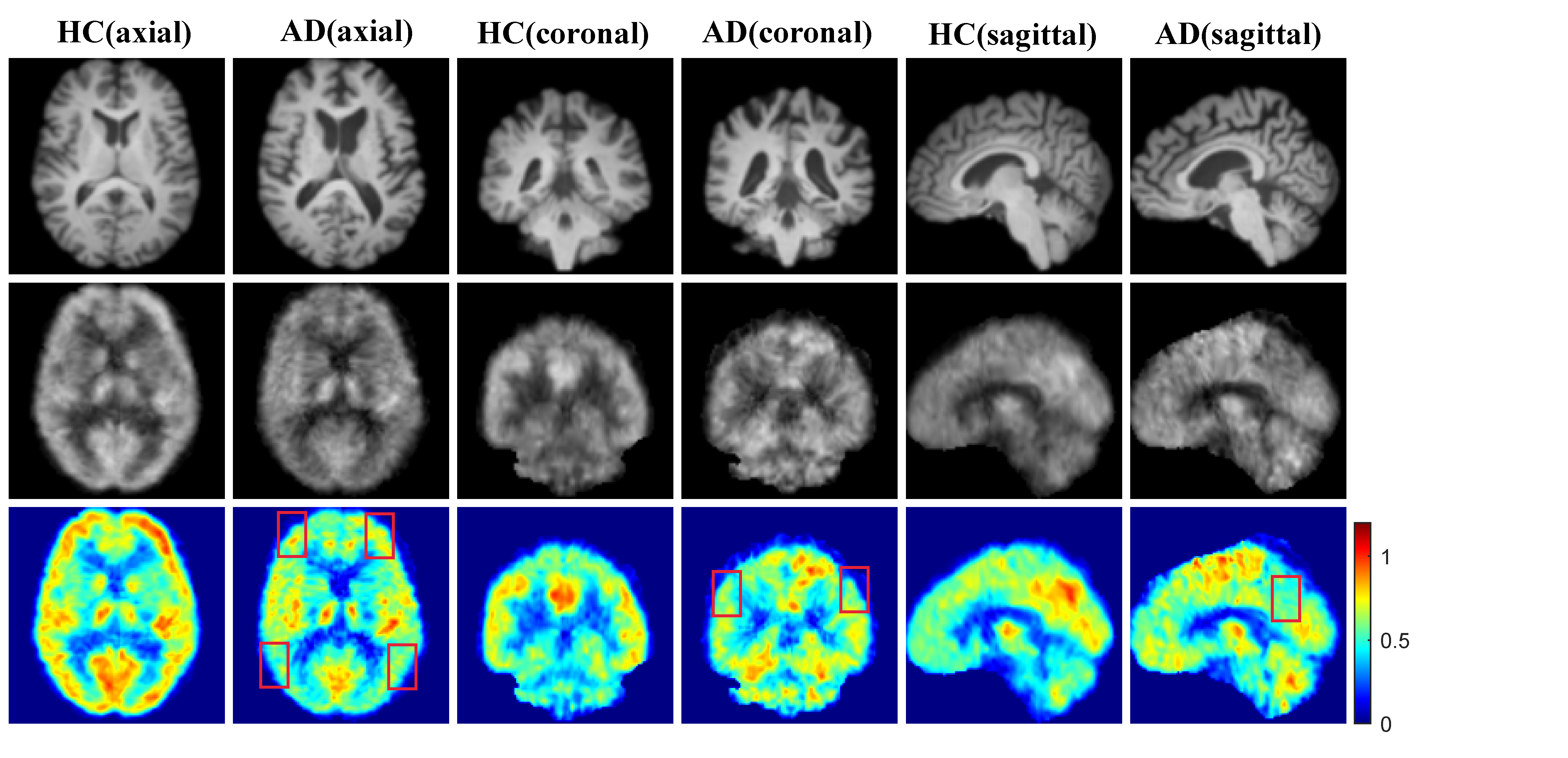}
    \caption{Synthetic PET from MRI for healthy controls (HC) and Alzheimer's disease (AD) patients in the axial plane, coronal plane, and sagittal plane. The first row shows MRI for HC and AD patients in the axial plane, coronal plane, and sagittal plane. The second row shows the PET, co-aligned with the MRI in the first row. The third row displays the synthetic PET from MRI in the first row by JDAM. The red rectangles were the regions of hypometabolism for Alzheimer's disease. }
    \label{fig:AD_CN}
\end{figure}
Figure \ref{fig:AD_CN} presents synthetic PET from MRI for both healthy controls and Alzheimer's disease patients in the axial plane, coronal plane, and sagittal plane. 
The frontal and parietal lobe hypometabolism is seen on PET imaging \cite{jack2018nia,shivamurthy2015brain}. The synthetic PET from MRI shows one of the regions highly associated with AD. The red rectangle and the black rectangle were marked after discussion with the physician.
The red rectangle shows hypometabolism in the regions of the frontal and occipital lobes in the axial. The red rectangle depicts decreased metabolism in the regions of the temporal lobe in the coronal. The red rectangle displays low metabolism in the regions of the parietal lobe in the sagittal. The experiment illustrates that synthetic PET by our method can be used as a reference for the diagnosis of AD.

\begin{figure}
    \centering
    \includegraphics[width=1\textwidth]{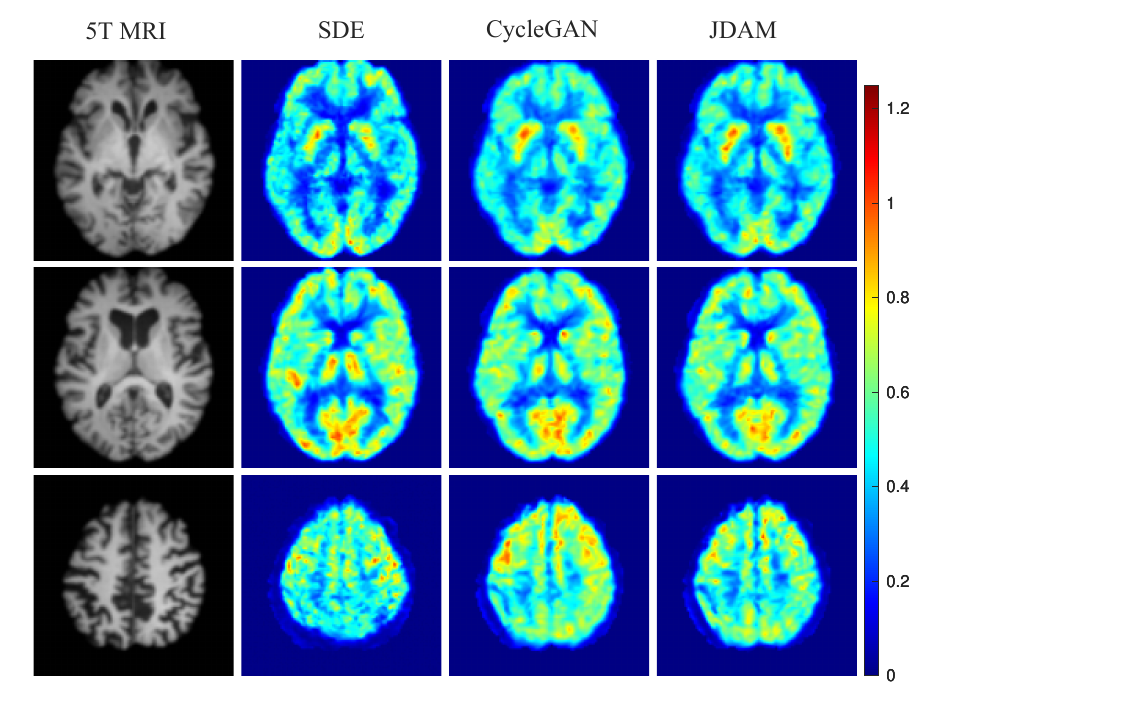}
    \caption{Synthetic PET from 5T MRI. The first row shows 5T MRI, along with the synthetic PET results generated by the score-based methods of SDE, CycleGAN, and JDAM in the upper layer of the axial plane. The second row shows images whose order is the same as the first row in the middle layer of the axial plane. The third row shows images whose order is the same as the first row in the lower layer of the axial plane.}
    \label{fig:5T}
\end{figure} 
\begin{figure}
    \centering
    \includegraphics[width=1\textwidth]{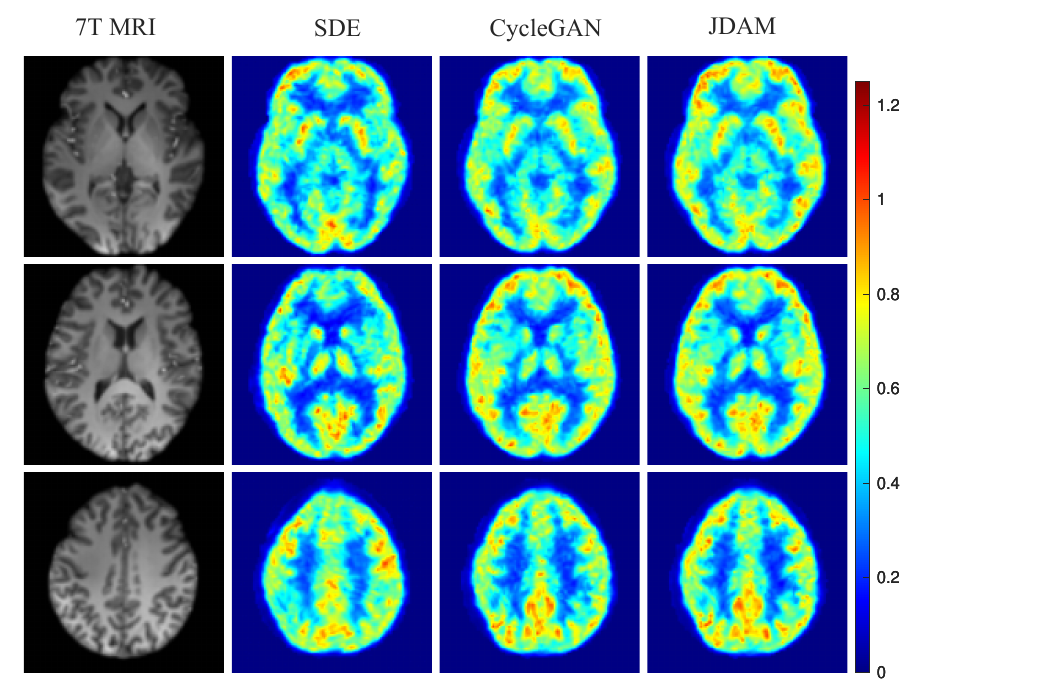}
    \caption{Synthetic PET using 7T MRI. The first row shows 7T MRI, along with the synthetic PET results generated by the score-based methods of SDE, CycleGAN, and JDAM in the upper layer of the axial plane. The second row shows images whose order is the same as the first row in the middle layer of the axial plane. The third row shows images whose order is the same as the first row in the lower layer of the axial plane.}
    \label{fig:7T}
\end{figure}
Based on the fact that field strength has no effect on PET imaging, JDAM was extended to ultra-high-field MRI. To evaluate the JDAM in the synthesizing PET corresponding to ultra-high-field MRI, an extensive set of experiments was conducted. Specifically, the trained model for the high-field MRI (3T MRI) was applied to the ultra-high-field MRI (5T MRI and 7T MRI). The 5T MRI was acquired by a 5T MRI scanner (uMR Jupiter, United Imaging, Shanghai, China). The synthesis PET images corresponding to 5T MRI are presented in Figure \ref{fig:5T}. The 7T MRI were acquired by a 7T MRI scanner (MAGNETOM Terra, Siemens Healthcare, Erlangen, Germany). All of the protocols were approved by our Institutional Reviews Board (IRB). The synthesis PET corresponding to 7T MRI is shown in Figure \ref{fig:7T}. Figure \ref{fig:5T} and Figure \ref{fig:7T} show the synthetic PET from the ultra-high-field MRI as guidance. The synthetic PET captures the background physiologic distribution of the PET image and has realistic spatial distributions. These results demonstrate the strong generalization ability of our JDAM model when extended to ultra-high-field strength. PET was generated from ultra-high-field MRI by SDE, CycleGAN, and JDAM. From a visual point of view, there are significant differences between the structure of the synthetic PET and the true MRI, even though the MRI has a higher resolution. The assumption is that PET and MRI have the same structure because PET and MRI are co-aligned. The synthesis PET using the ultra-high-field MRI was analyzed from two perspectives, including SNR and radiologist. On the one hand, the signal-to-noise ratio of the synthesis PET corresponding to the ultra-high-field MRI (5T MRI and 7T MRI) is higher than the SNR of the synthesis PET by the high-field MRI (3T MRI). The SNR of synthesis PET by JDAM is higher than the SNR of synthesis PET by CycleGAN from ultra-high-field MRI. On the other hand, the physician considers the results of the synthesis PET by JDAM and CycleGAN to be reasonable because the synthesis PET is symmetrical. From a physician's perspective, the symmetry of JDAM is better than CycleGAN. However, synthetic PET by SDE is not reasonable because it is asymmetrical, especially on the sagittal plane.

\section{Discussion}
\label{Sec:5}
In this study, we synthesized FDG PET from T1-weighted MRI using the joint probability distribution of the diffusion model at the high-field and the ultra-high-field. The joint probability distribution of the diffusion model effectively leverages MRI to learn the joint probability distribution of MRI and PET to enhance the quality of synthesized PET. Our results demonstrate that synthetic PET from high-field MRI and ultra-high-field MRI is possible. Synthetic PET captures the background physiologic distribution of PET imagery and has realistic spatial distributions. 
\begin{figure}
    \centering
    \includegraphics[width=1\textwidth]{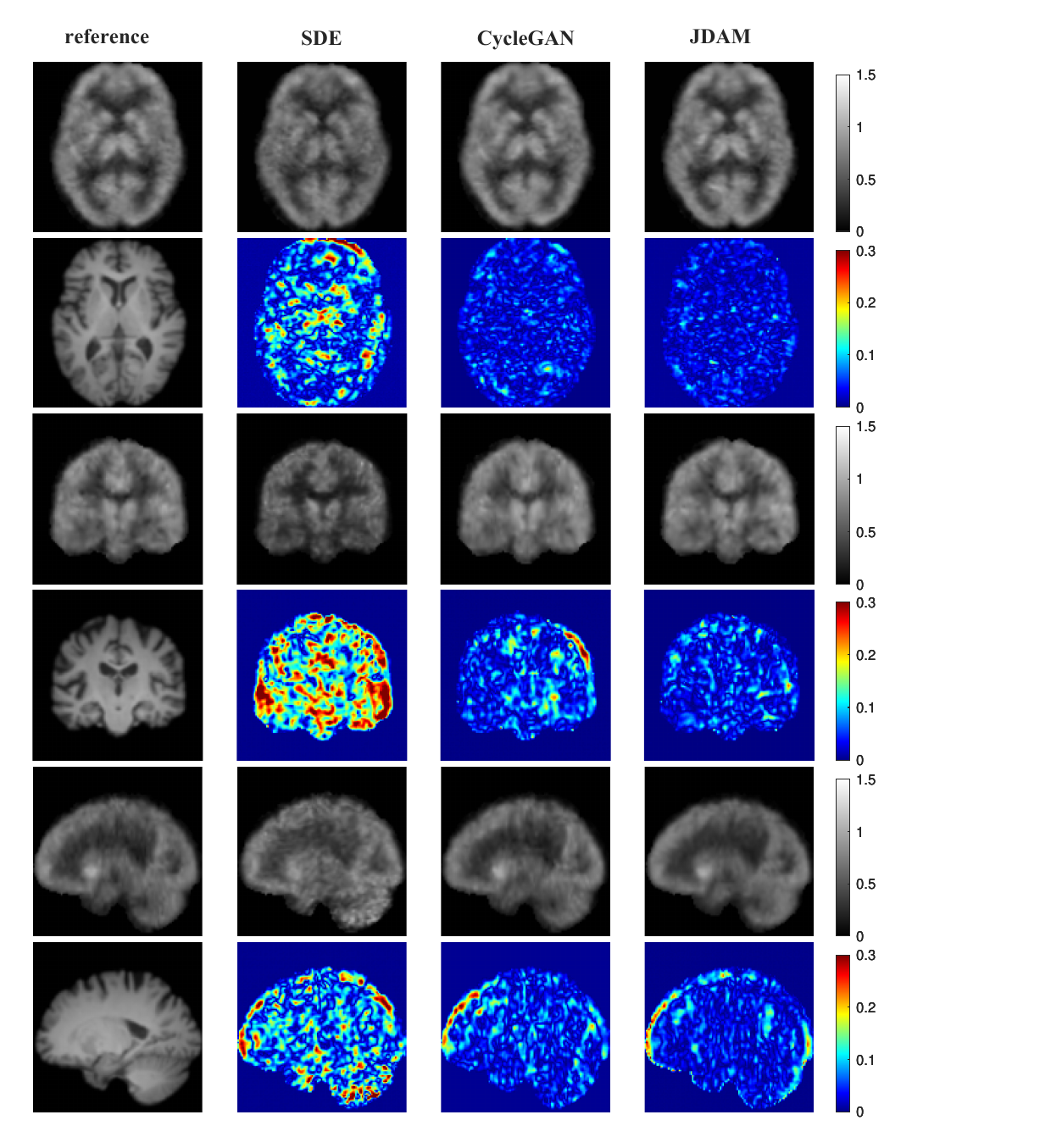}
    \caption{Error map of real PET and synthetic PET from high-field MRI. The first row is real PET in the axial plane and synthetic PET by SDE, CycleGAN, and JDAM. The second row shows the real MRI which is aligned with the real PET in the first row. The error map of the real PET and the synthetic PET by SDE, CycleGAN, and JDAM. The third row is real PET, synthetic PET in the coronal plane. The fourth row is real MRI, the error map of the synthetic PET and real PET in the coronal plane. The fifth row is real PET, synthetic PET in the sagittal plane. The sixth row is a real MRI in the sagittal plane, the error map of the synthetic PET and real PET in the sagittal plane.}
    \label{fig:error_map}
\end{figure}
In this study, we compared real PET and synthetic PET from high-field MRI by the SDE, CycleGAN, and JDAM. Figure \ref{fig:error_map} shows the error map of real PET and synthetic PET in the axial plane, coronal plane, and sagittal plane. The real PET and real MRI were co-aligned. PET was generated from MRI by SDE, CycleGAN, and JDAM. It can be observed that JDAM produced a high-quality PET image in which the error map shows an insignificant difference between the real and synthetic PET. The error of real PET and synthetic PET by JDAM is the smallest. However, synthetic PET is poor in SDE which shows a non-trivial discrepancy between real and synthetic PET. The absence of anatomical structure is probably the reason for the performance deterioration. In CycleGAN, PET synthesis performance was inadequate. The possible reason is that it cannot sample diversity.

In these comparison experiments, we confirmed the superiority of the JDAM over the SDE method and CycleGAN. However, some points still need discussion and improvement for JDAM. The accuracy of the SDE method is acceptable in theory, among all the results, its SSIM was the lowest. This is because the SDE approach synthesizes PET from noise, resulting in a synthetic PET image that differs from the real PET. 
Therefore, the SDE approach based on noise is not valid because it can not guarantee anatomical consistency between the MRI and the synthesis PET, which is essential for conducting research on both PET and MRI simultaneously. The JDAM was trained on the high-field MRI (3T MRI) and PET, and the training parameters from the high-field MRI were directly applied to the ultra-high-field MRI. 
Our concern is whether the high-field strength will have a strong effect on PET scanning. 
According to the literature \cite{sang2022mutual}, field strength does not have a significant effect on PET imaging. Synthetic PET from ultra-high-field MRI is referenced. Figure \ref{fig:5T} and Figure \ref{fig:7T} show the generated results of the PET that have three slices separately. The synthesis PET from ultra-high-field MRI (5T MRI and 7T MRI) was generated using the SDE, CycleGAN, and JDAM. JDAM can generate more details compared to the other two methods. This experiment validates the reliability of the JDAM for synthesizing PET from ultra-high-field MRI.
Similarly, synthetic PET by SDE is the worst because the structural similarity between MRI and PET cannot be guaranteed. CycleGAN and JDAM can guarantee structural similarity. The PSNR and SSIM of JDAM are higher than CycleGAN. JDAM can avoid the model crash that occurs for CycleGAN.

This work has two defects. Firstly, there was no way to compare the results due to the lack of PET in the ultra-high field. To solve the problem, the quality of synthetic PET was analyzed from the perspectives of the SNR and the radiologist. The SNR of synthetic PET from ultra-high-field MRI is higher than from high-field MRI. Likewise, the SNR of synthetic PET from 7T is higher than from 5T. Secondly, the attenuation correction of PET was not performed since the task of the paper is to generate PET by ultra-high-field MRI. In this paper, the quality of generation is considered primarily. A further study applies to MRI at the high field and the synthesis of PET at the high field for a variety of downstream tasks.

\section{Conclusions}
\label{Sec:6}
PET imaging is crucial for the diagnosis of brain diseases. However, due to its high cost and ionizing radiation, PET is not widely available. 
In this paper, synthetic PET from MRI at the high-field by JDAM model and extended to the ultra-high-field. Our method accurately captures the background physiologic distribution of PET imagery when using synthetic PET from ultra-high-field MRI as a guide. Additionally, the spatial distributions of PET synthesis images are realistic. It can be seen that the generation of corresponding PET by MRI at the high-field and the ultra-high-field is possible. 
The JDAM model not only improves the stability of the generation model but also generates PET from MRI accurately. The method shows high potential for cross-modal synthesis. In future research, accelerated imaging speed is one of the research directions.

\clearpage
\bibliography{refs}

\end{document}